\documentclass[12pt,a4paper]{article}
\usepackage{natbib}
\usepackage{graphicx}
\usepackage{amsmath}
\usepackage{amsfonts}
\usepackage{amssymb}
\usepackage{makeidx}
\usepackage{setspace}
\usepackage{geometry}
\usepackage{color}
\usepackage{hyperref}
\usepackage{xcolor}
\usepackage{epigraph}
\hypersetup{colorlinks, linkcolor={red!50!black}, citecolor={blue!50!black}, urlcolor={blue!80!black}}
\title{Recognition, recall, and retention of few-shot memories in large language models}
\author{
  Emin Orhan \\
  New York University\\
  \texttt{eo41@nyu.edu}
  }
\date{}

\begin{document}

\maketitle

\begin{abstract}
The training of modern large language models takes place in a regime where most training examples are seen only a few times by the model during the course of training. What does a model remember about such examples seen only a few times during training and how long does that memory persist in the face of continuous training with new examples? Here, we investigate these questions through simple recognition, recall, and retention experiments with large language models. In recognition experiments, we ask if the model can distinguish the seen example from a novel example; in recall experiments, we ask if the model can correctly recall the seen example when cued by a part of it; and in retention experiments, we periodically probe the model's memory for the original examples as the model is trained continuously with new examples. 

We find that a single exposure is generally sufficient for a model to achieve near perfect accuracy even in very challenging recognition experiments.~We estimate that the recognition performance of even small language models easily exceeds human recognition performance reported in similar experiments with humans (Shepard, 1967).~Achieving near perfect recall takes more exposures, but most models can achieve it in just 3 exposures. 

The flip side of this remarkable capacity for fast learning is that precise memories are quickly overwritten:~recall performance for the original examples drops precipitously over the first 10 training updates with new examples, followed by a more gradual decline.~Even after 100K training updates with new examples, however, some of the original examples are still recalled near perfectly.~A qualitatively similar retention pattern:~\textit{i.e.}~a fast decline in memory followed by a more gradual decline, with a more or less permanent memory ``permastore'' has been observed in human long-term memory retention studies before (Bahrick, 1984).~Finally, recognition is much more robust to interference than recall and memory for natural language sentences is generally superior to memory for stimuli without structure (sequences of random words or random characters).\footnote{All code, data, results related to this work are publicly available at \href{https://github.com/eminorhan/llm-memory}{this repository}.}
\end{abstract}

\section{Introduction}

\epigraph{\textit{Unless you have a rich reserve within, you can’t create anything.~That’s why I often say that creation comes from memory.~Memory is the source of your creation.~You can’t create something out of nothing.~Whether it’s from reading or from your own real-life experience, you can’t create unless you have something inside yourself.}}{--- Akira Kurosawa}

Modern language models are trained with vast amounts of data. The scale of the data ingested by these models during training is indeed so large that completing more than a single pass over the data is often computationally infeasible, especially for very large models. For example, GPT-3 was trained for 300B tokens, but the total size of its training data was roughly 500B tokens \citep{brown2020}. This means that each training example was effectively seen less than once on average by the model, although different subsets of the data were sampled at different frequencies, with parts of the data sampled up to 3.4 times during training while other parts sampled only 0.4 times. Even taking into account the possible repetition of parts of the data across the entire training set, it seems very likely that a large fraction of the training data was never seen more than a few times during training.

What exactly can a language model remember about training examples that it sees only a few times? How long does that memory persist as the model is continued to be trained with new examples? In this paper, we address these basic questions through simple recognition, recall, and retention experiments with language models inspired by experimental psychology. These questions are important both for their own sake as fundamental questions about the cognitive capabilities of modern language models and for their privacy and security implications.

\section{Related work}

This work builds on a sizeable recent literature investigating different aspects of memorization in language models \citep{carlini2019,carlini2021,carlini2022,zhang2021,kharitonov2021,mccoy2021,ippolito2022,tirumala2022,kandpal2022,mallen2022,haviv2022}. These studies generally emphasize the large capacity of language models to memorize chunks of text from their training data \citep{carlini2019,carlini2021,carlini2022,mccoy2021,ippolito2022,tirumala2022} with notable limitations in the case of ``long-tail'' examples that occur rarely throughout the training data \citep{kandpal2022,mallen2022}. Many of these studies, however, take the training data (and the training setup) of the pretrained models as given and hence do not control exposure. Our experiments, on the other hand, carefully control the models' exposure to the training examples whose memorization is studied, as explained in more detail below.

Some of these earlier studies also use formal measures of memorization that can be somewhat difficult to interpret intuitively \citep{carlini2019,tirumala2022}. By contrast, here we use easily interpretable measures of recognition and recall inspired by experimental studies of human memory. The recall measure we use here has been used in earlier studies \citep{carlini2022,mccoy2021,haviv2022}, but \textit{verbatim} recall, on its own, can be insensitive to weaker, more subtle forms of memory, hence it can be too strict as a measure of memory, \textit{cf.}~\cite{ippolito2022}. As we show below, our recognition measure can detect more subtle forms of memory that may be undetectable with \textit{verbatim} recall through greedy decoding. Recognition tests also allow us to directly compare memory in language models with human memory performance measured in similar experiments \citep{shepard1967}. 

Earlier studies generally do not consider the retention or durability of memories once these memories are inducted into the model. The only exception we know of is \cite{tirumala2022}, which, as mentioned above, used more formal measures of memorization. Here, we complement this study by tracking the retention of memories through more easily interpretable recognition and recall measures. 



\section{Experiments}
\label{expt_sec}
\begin{figure}
  \centering
    \includegraphics[width=1.0\textwidth, trim=0mm 0mm 0mm 0mm, clip]{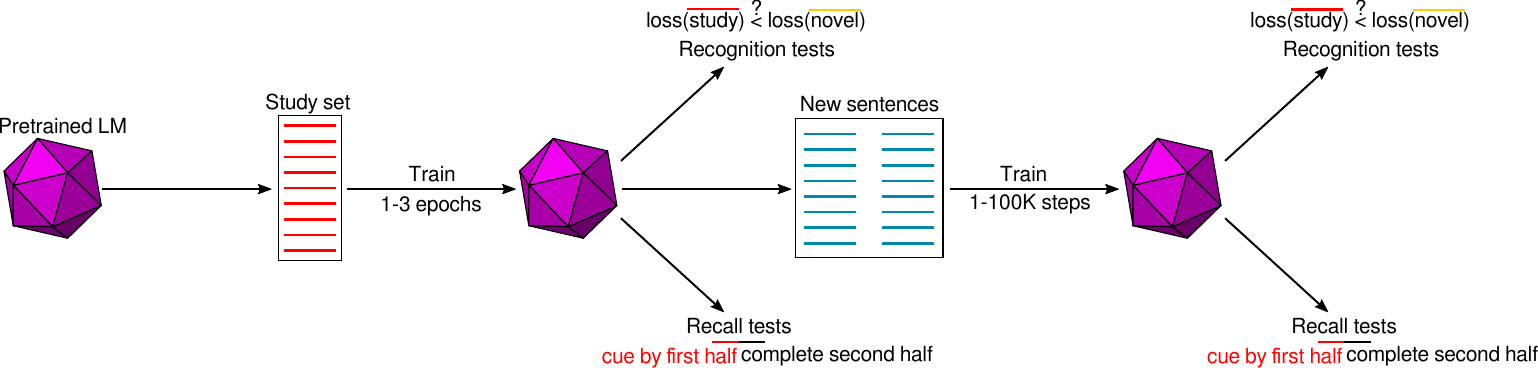}
  \caption{Schematic overview of the experiments. See Section~\ref{expt_sec} for a detailed description of the experiments.}
  \label{schema_fig}
\end{figure}

The general setup of our experiments is as follows (see Figure~\ref{schema_fig} for a schematic overview of the experiments).~We start with a pretrained language model.~We train it with a set of study sentences it has never seen before (which can be different in different experiments). Each study sentence is presented only 1-3 times to the model, \textit{i.e.} 1-3 epochs of training over the study set. We then probe the model's memory for the study sentences by giving it a recognition or a recall test. Recall tests are generally very stringent tests of memory, hence here we complement them with more sensitive recognition tests that can detect more subtle forms of memory that may not be accessible to recall. 

In the recognition tests, the model is asked to tell apart the study sentences from unseen or novel foils. In the recall tests, the model is cued (or prompted) with part of a study sentence and is asked to complete the remainder of the sentence. In retention experiments, after the study phase, we keep on training the model with new sentences and perform the recognition and recall tests for the original study set at regular intervals. This tells us how fast the memory for sentences in the original study set deteriorates in the face of continuous training of the model with new sentences.

For each experiment, we perform hyperparameter searches over the learning rate and batch size to maximize memory performance. The results of all our hyperparameter runs (approximately 20K runs in total) are publicly available from \href{https://huggingface.co/datasets/eminorhan/llm-memory}{this} Huggingface Datasets repository. In general, we find that our results are not too sensitive to the hyperparameter choices: \textit{i.e.}~there is a wide range of learning rates that lead to similar memory performance if appropriate adjustments are made to the batch size (and \textit{vice versa}).

\subsection{Recognition experiments}
In the recognition experiments, after the model is trained on the study set, we test whether it can distinguish the studied items from unseen or novel foils. In each trial, we present the model with a study item and a foil. We then compare the losses assigned by the model to the study item \textit{vs.}~the foil. We use these language modeling losses as a proxy for familiarity, with lower losses corresponding to more familiar stimuli. So, if the model assigns a lower loss to the study item: \textit{i.e.}~$loss(\mathbf{x}_s)<loss(\mathbf{x}_f)$, where $\mathbf{x}_s$ denotes the study sentence and $\mathbf{x}_f$ denotes the foil, we consider the model's response for that trial correct, otherwise the response is considered incorrect.

Note that by changing the study set, we can investigate how well the model can remember different types of stimuli (\textit{e.g.} well-formed, meaningful sentences \textit{vs.} length-matched random strings) and by changing the similarity between the study items and the foils, we can probe the granularity of the model's memory for the study items. We will employ both types of manipulations in the experiments below.

\subsubsection*{Experiment 1: normal sentences vs.~normal sentences}
We first curated a dataset of 6007 English sentences collected from four blogs on Substack focusing on a variety of topics (blogs by \href{https://rychappell.substack.com}{Richard Chappell}, \href{https://astralcodexten.substack.com/}{Scott Alexander}, \href{https://fakenous.substack.com/}{Michael Huemer}, and \href{https://betonit.substack.com/}{Bryan Caplan}). To make sure these sentences do not appear among the training data of any of the pretrained language models we use in the following experiments, we only considered relatively recent posts published between April-October 2022. The complete set of sentences can be found \href{https://github.com/eminorhan/llm-memory/blob/master/data/substack_all.json}{here}.

For this first experiment, we randomly sampled 600 study sentences and 600 separate foils from this dataset. The number 600 was chosen based on \cite{shepard1967}, who ran a similar sentence recognition memory experiment with humans. One of our goals for this experiment is to compare sentence recognition memory in humans vs.~language models. Although we do not have access to the exact sentences used by \cite{shepard1967} in his sentence recognition memory experiment, the design of our experiment otherwise exactly replicates the design of his experiment, so we believe it is reasonable to directly compare the human performance estimate from \cite{shepard1967} and the language model estimates from our experiment against each other.

We replicated this experiment (as well as all other experiments below) four times, sampling different study and foil sets each time. Note that randomization helps prevent any systematic differences between the study and foil items. An example study-foil pair in this experiment would be as follows:
\begin{itemize}
\item \textit{Study:} Don’t I specifically believe in things many people have found self-evidently absurd (eg the multiverse, AI risk)?
\item \textit{Foil:} This is where you go to read about Shakespeare, post-modernism, arthouse films, and Chinese tapestries, right?
\end{itemize}

\subsubsection*{Experiment 2: normal sentences vs.~their paraphrases}
Because the study items and foils are different sentences in the previous experiment, a model can, in principle, perform well even with relatively coarse-grained memories of the study sentences (\textit{e.g.}~by remembering only the semantic gist of each sentence).~To probe the granularity of the memory, in the second experiment, we used paraphrases of the study sentences as foils in each trial.~For example, a study-foil pair would be as follows:
\begin{itemize}
\item \textit{Study:} And while the slogan horrifies people who oppose immigration, the undecided are usually mildly intrigued.
\item \textit{Foil:} People who oppose immigration are usually horrified by the slogan, while people who are undecided are usually mildly interested.
\end{itemize}

Thus, a relatively coarse-grained memory would not be able to distinguish between the study item and the foil in this experiment. We generated the paraphrases of the study sentences using the \texttt{text-davinci-002} model on the OpenAI API (the Python code, including the prompt, used for generating the paraphrases can be found \href{https://github.com/eminorhan/llm-memory/blob/master/utils/create_paraphrases.py}{here}).

\subsubsection*{Experiment 3: normal sentences vs.~single word substitutes}
In Experiment 3, we probe the granularity of the model's memory further by making the foils even more similar to the corresponding study items. More specifically, we generate each foil by making a single word synonym substitution in the corresponding study sentence. An example study-foil pair would thus be as follows:
\begin{itemize}
\item \textit{Study:} The other people start out with money that belongs to them and that they have no obligation to give to you.
\item \textit{Foil:} The other people start out with money that belongs to them and that they have no duty to give to you.
\end{itemize}

Here the word \textit{obligation} in the study sentence is replaced with the synonymous \textit{duty}. We again generate the foils using the \texttt{text-davinci-002} model on the OpenAI API (the Python code used for generating the foils can be found \href{https://github.com/eminorhan/llm-memory/blob/master/utils/create_synonyms.py}{here}).

\subsubsection*{Experiment 4: random words vs.~random words}
To investigate the extent to which the model's memory depends on the rich structural and semantic regularities in well-formed meaningful sentences of natural language, we next tried alternative study stimuli lacking these regularities. In Experiment 4, we used sequences of 25 random words  sampled uniformly from the collection of words appearing in the novel \textit{Moby Dick} (a vocabulary of roughly 19k words in total). We generated both study items and foils in this way, so an example study-foil pair from this experiment might look like the following:
\begin{itemize}
\item \textit{Study:} eating presentations sons sounds ahaz froissart violence placidly gaffman popularize crying swiftest continually exceeding nowhere honours descriptively erromanggoans unrecorded lathering octher diameter honeycombs perchance java
\item \textit{Foil:} caw swiftly describe reverie carrol directed orientals opposite sentiment drums pilau fry exclaimed idle sleeves wintry son immeasurable absorbingly suffering innocent gaspings rokovoko embark sibbald
\end{itemize}

\subsubsection*{Experiment 5: random words vs.~normal sentences}
Pretrained language models generally assign a higher prior probability to well-formed, meaningful natural language sentences than to sequences of random words. In Experiment 5, we sought to determine whether 1-3 exposures to random word sequences during the study phase would be sufficient to reverse this preference and make them more likely than unseen normal natural language sentences. Therefore, this experiment used sequences of random words \textit{vs.}~natural language sentences as study-foil pairs. An example study-foil pair in this experiment would thus be as follows:
\begin{itemize}
\item \textit{Study:} event scrambled wonst vessel maelstrom necessitated hurt insinuating boiler glided gravity spades dragged swayed colourless plight swing bisons slave cloistered ramadans grappling supplemental effaced outfit
\item \textit{Foil:} This loneliness synergizes with Leto's immense boredom, an ennui enhanced by experience and near-omniscience.
\end{itemize}

\subsubsection*{Experiment 6: random strings vs.~normal sentences}
Experiment 6 is a simple variation on the previous experiment, where we remove the within-word structure in study items as well. We do this by starting with a normal sentence, shuffling the words within the sentence, and then shuffling the characters within each word. This produces a random string that lacks any regularities at the sentence level or at the word level. We use these random strings as study items and the normal sentences they were generated from as the corresponding foils. An example study-foil pair from this experiment might thus look like this:
\begin{itemize}
\item \textit{Study:} eptyrt atth humc hte rwe,oveH lewl lasw nkows ciemr of eud isht cetfyrple solaic si ceicsen drug sle.hetsvme in to ostm nevyereo
\item \textit{Foil:} However, pretty much everyone in social science knows perfectly well that most of this crime is due to the drug laws themselves.
\end{itemize}

\subsection{Recall experiments}
In the recall experiments, after the language model is trained on the study sentences during the study phase, we cue (or prompt) the model with the first half of each study sentence (after tokenization) and ask the model to complete the second half of the sentence. For convenience and simplicity, we use greedy decoding to generate the model completions. We measure the performance of the model by computing the Rouge-L score, \textit{i.e.} the length of the longest matching substring between the ground truth sentence and the prompt + model generated completion \citep{lin2004}. If there is no match between the ground truth completion and the model completion, we thus expect a Rouge-L score of roughly 0.5, whereas a Rouge-L score of 1 indicates perfect recall.

We perform these recall experiments for each of the three types of stimuli used in the recognition experiments above: namely, (i) normal well-formed meaningful natural language sentences, (ii) sequences of random words, and (iii) random strings.

\subsection{Retention experiments}
In the retention experiments, after the model is trained on the study set, we keep training it on a separate set of novel sentences. We use sentences from the CNN/Daily Mail dataset for this purpose. We combine all training, validation, and test data from this dataset for a total of 224446 sentences. We perform the recognition and recall tests described above (with the original study set) at regular intervals, while the model is being trained on the new sentences from the CNN/Daily Mail dataset. The goal of these retention experiments is to find out how fast the memory for the original study sentences deteriorates with continuous training of the model with new sentences. Due to computational constraints, we perform the retention experiments only with our best overall model, namely the \texttt{gpt-j-6B} model trained for 3 epochs over the study set (3 exposures).\footnote{Note that the CNN/Daily Mail dataset used in our retention experiments was not in the training set of \texttt{gpt-j-6B}, therefore sentences from this dataset are novel for the model.}

\section{Models}
To ensure that our results are robust to variations in architecture, pretraining data, and other pretraining details, we considered four different classes of causal language models: BLOOM, OPT, GPT-2, and GPT-Neo. To investigate the effect of model size, we considered models of different sizes for each class, up to 13B parameter models (we use the official model names from the Huggingface \texttt{transformers} library below): 
\begin{itemize}
    \item For BLOOM: \texttt{bloom-560m}, \texttt{bloom-1b1}, \texttt{bloom-1b7}, \texttt{bloom-3b}, \texttt{bloom-7b1}
    \item For OPT: \texttt{opt-125m}, \texttt{opt-350m}, \texttt{opt-1.3b}, \texttt{opt-2.7b}, \texttt{opt-6.7b}, \texttt{opt-13b}
    \item For GPT-2: \texttt{gpt2}, \texttt{gpt2-medium}, \texttt{gpt2-large}, \texttt{gpt2-xl}
    \item For GPT-Neo: \texttt{gpt-neo-125m}, \texttt{gpt-neo-1.3B}, \texttt{gpt-neo-2.7B}, \texttt{gpt-j-6B}
\end{itemize}
for a total of 19 distinct models varying in size from 125M parameters up to 13B parameters, covering a range that is roughly two orders of magnitude in size. Note that we included \texttt{gpt-j-6B} in the GPT-Neo class above. Models with more than 6B parameters (\texttt{bloom-7b1}, \texttt{opt-6.7b}, \texttt{opt-13b}, \texttt{gpt-j-6B}) were trained on four A100 GPUs (with 80GB GPU memory) using fully sharded data parallelism (FSDP), with CPU offloading of the parameters in the case of the largest model (\texttt{opt-13b}). The remaining models were all trained on single A100 GPUs.

\section{Results}

\subsection{Recognition experiments}

\begin{figure}
  \centering
    \includegraphics[width=0.9\textwidth, trim=0mm 0mm 0mm 0mm, clip]{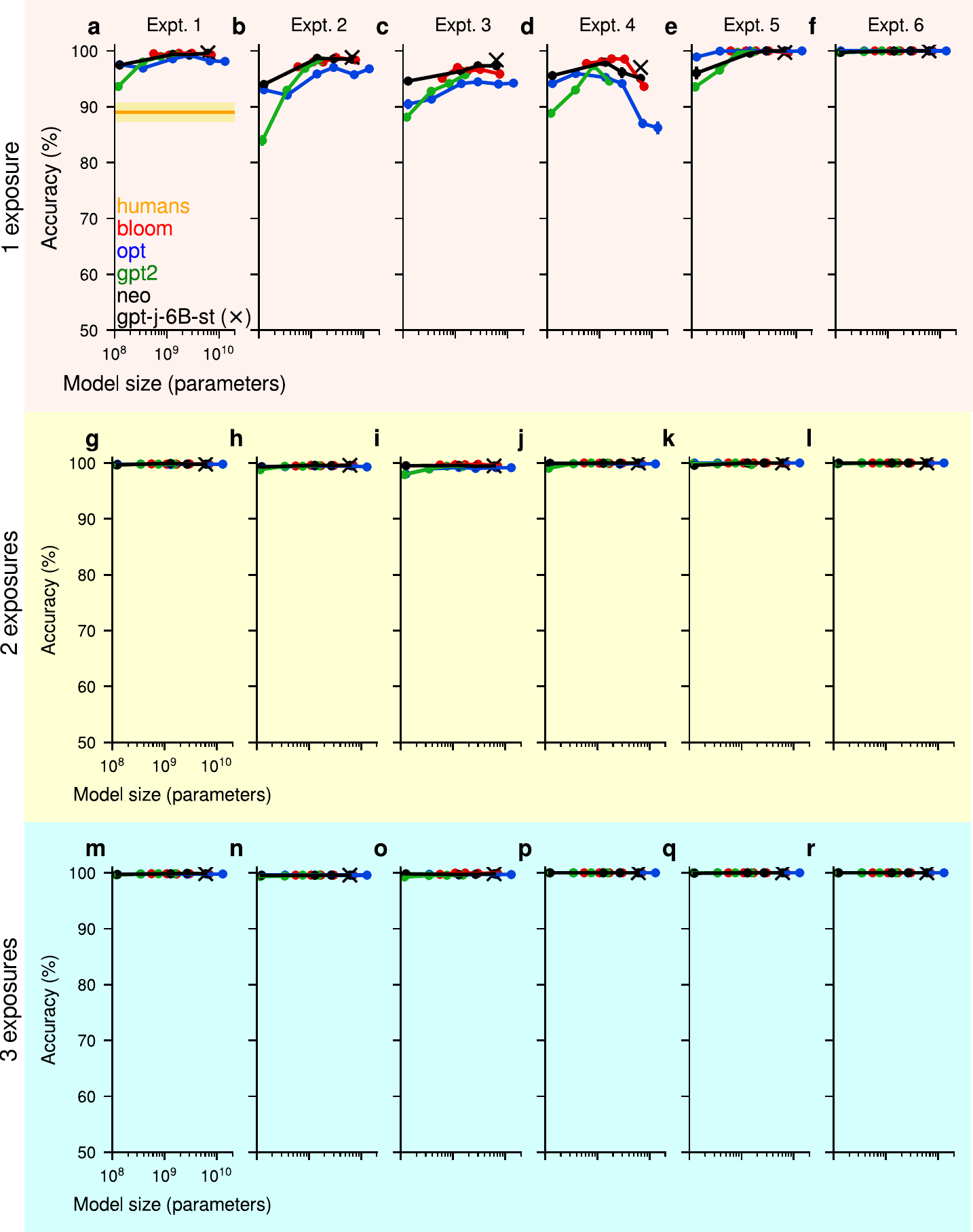}
  \caption{Results of all recognition experiments. The top row shows the results for the 1-exposure condition, where the models see the study sentences only once; the middle row shows the results for the 2-exposure condition (\textit{i.e.} 2 epochs of training over the study set); the bottom row shows the results for the 3-exposure condition. Different colors represent different model classes. Error bars (too small to see in most cases) represent standard errors over 4 replications of each experiment.}
  \label{recognition_fig}
\end{figure}

Figure~\ref{recognition_fig} shows the results of all our recognition experiments. Almost all models achieve very high recognition accuracies even in the most challenging single exposure condition (1 epoch of training over the study set) in all six experiments (top row in Figure~\ref{recognition_fig}). As expected, models generally find it harder to distinguish study sentences from their paraphrases or single-word substitutes (Experiments 2-3) than to distinguish them from completely unrelated novel sentences (Experiment 1), but recognition accuracy remains very high in all experiments.

In Figure~\ref{recognition_fig}a, we compare the model recognition memory for natural language sentences with human recognition memory estimated in a similar experiment by \cite{shepard1967}. As noted above, although we do not have the exact natural language sentences used in this classic study, the single exposure condition of our Experiment 1 is otherwise very similar to the sentence recognition memory experiment reported in \cite{shepard1967}, hence we expect the human recognition memory performance estimated in that paper to be a reasonably good estimate for our Experiment 1 as well.\footnote{The sentences used in \cite{shepard1967} were shorter and simpler than the ones used in our Experiment 1, so it is likely that the sentence recognition memory experiment in \cite{shepard1967} was, if anything, easier than our experiment.} Figure~\ref{recognition_fig}a clearly shows that the recognition accuracy of all models easily exceeds the estimated human recognition accuracy. 

At two and three exposures, all models achieve near perfect recognition accuracy in all experiments (middle and bottom rows in Figure~\ref{recognition_fig}). These results suggest that gradient descent may be remarkably efficient in ingesting information into a model. 

\begin{figure}
  \centering
    \includegraphics[width=0.9\textwidth, trim=0mm 0mm 0mm 0mm, clip]{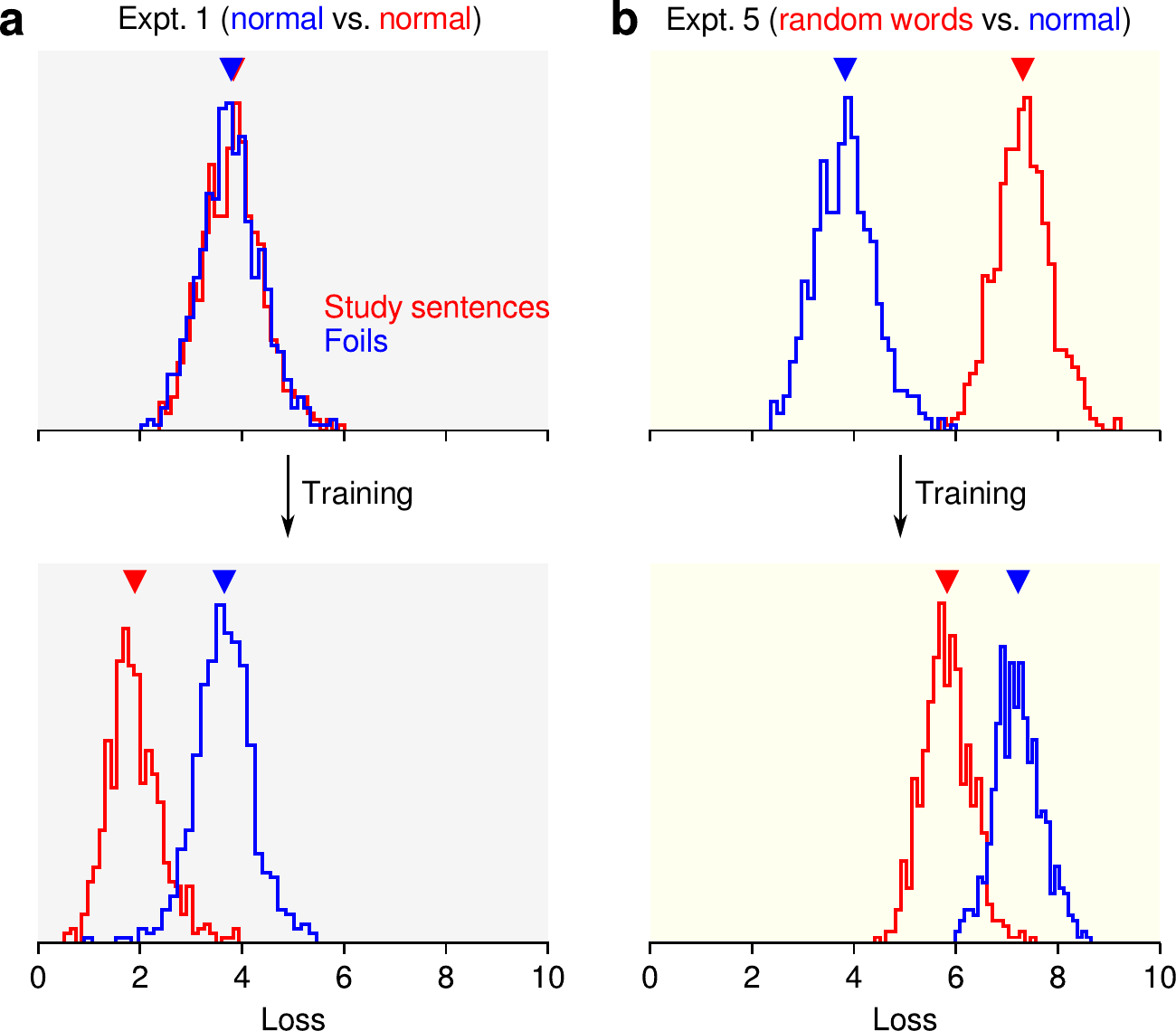}
  \caption{Loss distributions for the study sentences and foils in Experiment 1 (a) and in Experiment 5 (b), before and after a single exposure to the study sentences (top and bottom rows, respectively). In Experiment 1, where both study sentences and foils are normal natural language sentences, exposure to the study sentences increases their likelihood (\textit{i.e.} decreases their loss), but does not change the loss distribution of the foils. In Experiment 5, on the other hand, exposure to random word sequences increases their likelihood and decreases the likelihood of the normal sentences at the same time. The results shown here are from the \texttt{bloom-7b1} model. Inverted triangles indicate the means of the distributions.}
  \label{recognition_loss_hist_fig}
\end{figure}

It might be expected that models pretrained with natural language sentences should be good at few-shot recognition of natural language sentences (Experiments 1-3), but the models are also very good at few-shot recognition of sequences of random words or random strings (Experiments 4-6). In one of our more surprising results (Experiment 5), we observe that a single exposure to a set of random word sequences is enough to reverse the models' initial preference for normal natural language sentences over such random word sequences. Mechanistically, we find that exposure to random word sequences increases the likelihood of such stimuli under the model, while simultaneously decreasing the likelihood of the normal natural language sentences (Figure~\ref{recognition_loss_hist_fig}b).\footnote{Note that this does not necessarily mean that the models' knowledge of normal natural language sentences is irreversibly lost. That knowledge may be quickly recoverable after a brief retraining of the model with normal sentences again. We leave a more detailed investigation of this possibility to future work.} In contrast, in Experiment 1, where both study sentences and foils are normal natural language sentences, exposure to the study set increases the likelihood of the study sentences without affecting the likelihoods of the foils (Figure~\ref{recognition_loss_hist_fig}a). 

\subsection{Recall experiments}

\begin{figure}
  \centering
    \includegraphics[width=0.9\textwidth, trim=0mm 0mm 0mm 0mm, clip]{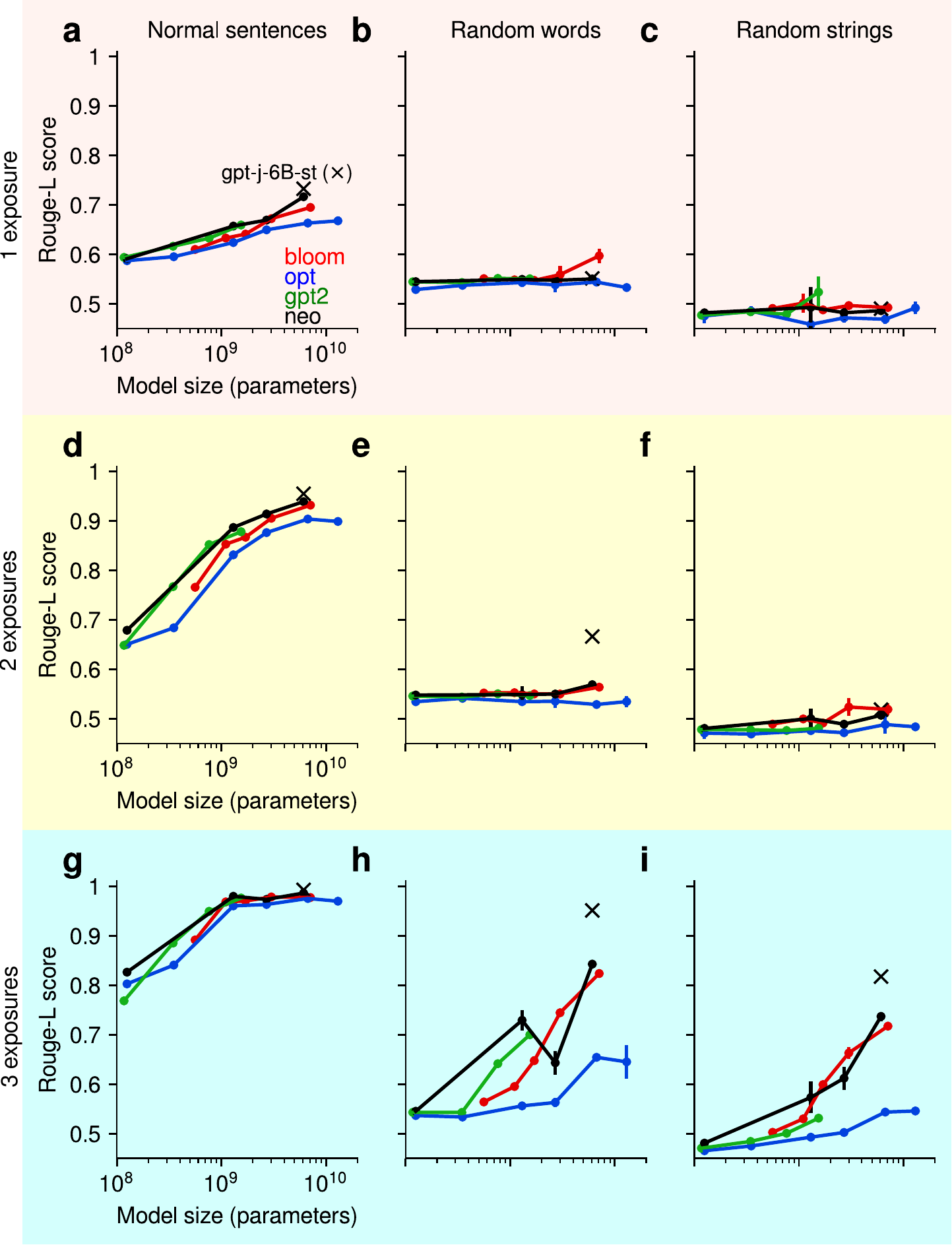}
  \caption{Results of all recall experiments. As in Figure~\ref{recognition_fig}, the top row shows the results for the 1-exposure condition; the middle row shows the results for the 2-exposure condition; and the bottom row shows the results for the 3-exposure condition. Different colors represent different model classes. The three columns show the recall performance for normal natural language sentences, random word sequences, and random strings, respectively. Error bars represent standard errors over 4 replications of each experiment.}
  \label{recall_fig}
\end{figure}

Figure~\ref{recall_fig} shows the results of all our recall experiments. With 1 exposure only to the study sentences (top row in Figure~\ref{recall_fig}), the models cannot achieve anywhere near perfect recall for any of the three types of stimuli. With 2 exposures (middle row in Figure~\ref{recall_fig}), the best models (typically the larger models) are able to achieve very high recall performance for the normal natural language sentences, but they are still near chance levels for the other types of stimuli (random word sequences and random strings). With 3 exposures (bottom row in Figure~\ref{recall_fig}), most models are able to achieve near perfect recall for the normal natural language sentences and the best models achieve high recall accuracy even for random word sequences and random strings.

\begin{figure}
  \centering
    \includegraphics[width=0.9\textwidth, trim=0mm 0mm 0mm 0mm, clip]{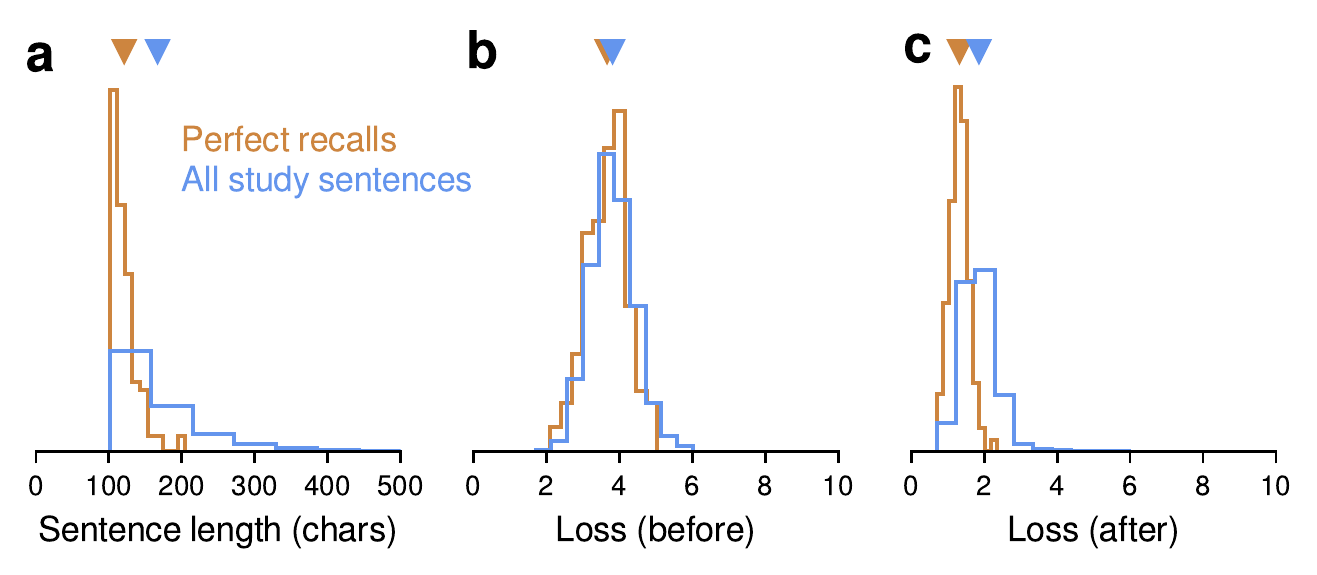}
  \caption{What distinguishes perfectly recalled sentences from other study sentences? The results shown here are from the \texttt{bloom-7b1} model after a single exposure to normal natural language sentences. Inverted triangles indicate the means of the distributions.}
  \label{perfect_recall_fig}
\end{figure}

We next investigate what distinguishes a perfectly recalled study sentence from other study sentences. To this end, we identified the normal natural language sentences perfectly recalled by the \texttt{bloom-7b1} model after a single exposure and compared them with all the other sentences in the same study set. Perfectly recalled study sentences were, on average, shorter than the other study sentences (Figure~\ref{perfect_recall_fig}a). This is to be expected, since perfectly recalling a shorter sentence is easier than perfectly recalling a longer sentence. There was no difference between the likelihoods of the perfectly recalled study sentences vs.~all other study sentences under the model before the model was exposed to the study set (Figure~\ref{perfect_recall_fig}b). This suggests that any features that may distinguish the perfectly recalled sentences from the other sentences are likely related to more subtle aspects of these sentences than their overall prior probability under the model. As expected, after exposure to the study set, the perfectly recalled sentences, on average, have a higher likelihood (lower loss) under the model than the other study sentences (Figure~\ref{perfect_recall_fig}c). 

\subsection{Retention experiments}

\begin{figure}
  \centering
    \includegraphics[width=1.0\textwidth, trim=0mm 0mm 0mm 0mm, clip]{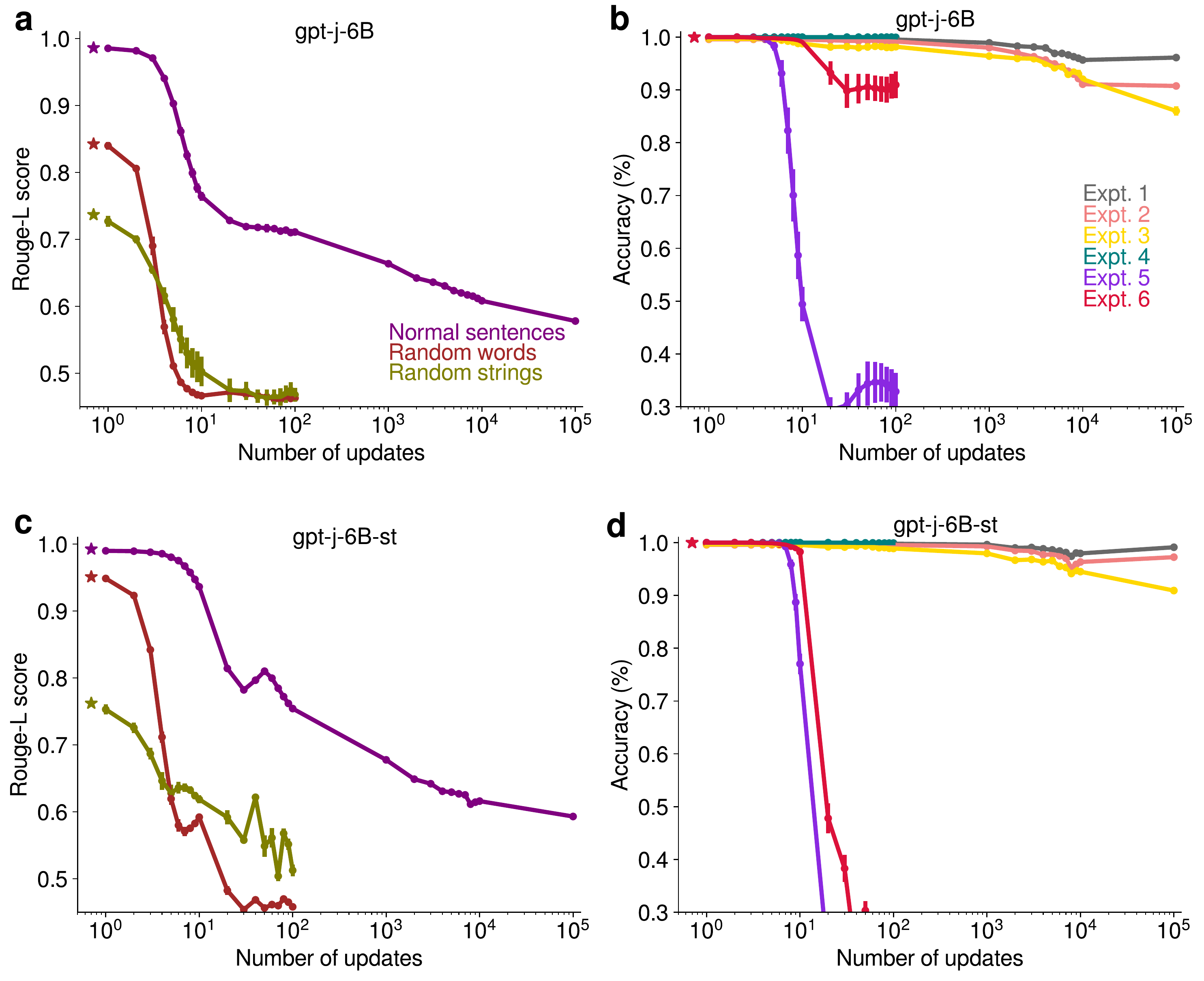}
  \caption{Results of all retention experiments. Error bars represent standard errors over 4 independent replications of each experiment. (a) Recall and (b) recognition performance of the \texttt{gpt-j-6B} model for the original study sentences as the model is continuously trained with new sentences for 1-100K iterations (\textit{x}-axis). (c) and (d) are similar, but for the \texttt{gpt-j-6B-st} model. The star signs ($\star$) indicate the performance of the models before the beginning of the retention experiment, \textit{i.e.} before the model is trained with any new sentences.}
  \label{retention_fig}
\end{figure}

Figure~\ref{retention_fig} shows the results of our retention experiments. In general, recall performance for the original study set drops precipitously over the first 10 or so training updates as the model (here \texttt{gpt-j-6B}) is continually trained with new sentences (Figure~\ref{retention_fig}a). With the normal natural language sentences, this is followed by a more gradual decline and even after a very large number of training updates (100K updates), recall performance remains above chance (purple in Figure~\ref{retention_fig}a). With the other types of stimuli (random word sequences and random strings), on the other hand, recall performance already drops to chance levels after around 10 updates. 

We find the recognition performance to be generally much more robust to interference from continual training with new sentences (Figure~\ref{retention_fig}b). The only exceptions are the recognition experiments that involve distinguishing studied random word sequences or random strings from novel normal natural language sentences (Experiments 5-6), where the recognition performance drops more sharply. This is not unexpected, since the model is being trained with normal natural language sentences (from the CNN/Daily Mail corpus) in this phase, so the model increases the likelihood of such sentences at the expense of more structureless stimuli such as random word sequences or random strings, even though these structureless stimuli were previously memorized by the model. Interestingly, however, the model has no difficulty distinguishing studied random word sequences from unstudied random word sequences (Experiment 4 in Figure~\ref{retention_fig}b), suggesting that it still assigns a higher likelihood to the studied ones compared to the unstudied ones, hence it retains a memory of those sequences even though the memory is not strong enough to enable successful recall.

\subsection{Are the results with fully pretrained models representative of what happens during pretraining?}
\label{st_sec}
The experiments so far considered only fully pretrained language models. A natural concern regarding these experiments is thus whether the results obtained with such fully pretrained models are representative of what happens \textit{during} pretrainig. To address this concern, we replicated all our experiments with a \texttt{gpt-j-6B} model trained from scratch on the \texttt{wikitext-103} dataset\footnote{Available from Huggingface Datasets as \href{https://huggingface.co/datasets/wikitext}{\texttt{wikitext-103-raw-v1}}.} for a total of 143M tokens only, which is a tiny fraction (0.04\%, to be precise) of the 402B tokens the original \texttt{gpt-j-6B} model was pretrained for.\footnote{We note, however, that our pretraining setup is different from the original \texttt{gpt-j-6B} pretraining run: we use a batch size of 16 and a truncated context length of 128 and train the model for exactly 70K iterations, compared with a context length of 2048 and 385K training iterations for the original \texttt{gpt-j-6B} pretraining run.} We call this model \texttt{gpt-j-6B-st} (\texttt{st} stands for \textit{short training}). The results of all recognition and recall experiments with this model are shown by the cross symbols ($\times$) in Figures~\ref{recognition_fig} and \ref{recall_fig}, respectively. 

In general, \texttt{gpt-j-6B-st} performs very similarly to \texttt{gpt-j-6B} in most experiments. The only difference is that \texttt{gpt-j-6B-st} performs slightly better than \texttt{gpt-j-6B} in some of the recall experiments with structureless stimuli, \textit{i.e.} random word sequences or random strings (Figure~\ref{recall_fig}e, h, and i). This could be because, having been trained on a lot more natural language data than \texttt{gpt-j-6B-st}, \texttt{gpt-j-6B} might have a stronger preference or inductive bias for natural language stimuli, as a result hurting its ability to remember more structureless stimuli.

\texttt{gpt-j-6B-st} behaves qualitatively very similarly to \texttt{gpt-j-6B} in the retention experiments as well (Figure~\ref{retention_fig}c-d). We thus conclude that all recognition, recall, and retention results reported in this work are likely representative of the memory properties of a language model through a very substantial chunk of its pretraining run.

\section{Limitations}
In this section, we briefly note two main limitations of the current work:
\begin{itemize}
    \item \textbf{Need to evaluate bigger, more capable models:} The largest language model considered in this work is a 13B parameter model (\texttt{opt-13b}). Today's best language models are typically larger and more capable than the models evaluated here. It remains to be seen whether or to what extent the memory properties of these bigger, more capable models differ from the memory properties of the models considered here. The models considered in this work are also all trained with the standard language modeling objective only. Modern language models, on the other hand, are often finetuned to follow instructions based on human feedback \citep{ouyang2022} in order to make them more controllable and more useful to end users. It remains unclear whether or how such ``instruction tuning'' might affect the memory properties of language models. 
    \item \textbf{Limited experimental probes of memory:} We have used two different methods to probe memory in language models: recognition tests and conditional recall tests with greedy decoding. Recognition is generally a less stringent test of memory than recall. By varying the similarity between the study items and the foils in our recognition tests (as we did), we can measure the granularity of the memories to some extent, but recognition experiments by themselves cannot tell us what exactly the models remember about the studied sentences or how exactly they are able to do the recognition tests (\textit{e.g.} what features they rely on to distinguish the studied items from foils). On the other hand, exact recall with greedy decoding is probably too stringent as a measure of memory. We thus need more fine-grained and more informative measures of memory. A very sensible idea in this direction is to use sampling for conditional generation (at different temperatures), instead of greedy decoding. This would presumably give us a much more detailed view of the model's ``memory landscape'' around a particular study item.
\end{itemize}

\section{Discussion}
Our results suggest that large language models are remarkably effective at incorporating a new piece of information into their parameters through gradient descent. The head-to-head comparison with human recognition memory performance in Figure~\ref{recognition_fig}a suggests that even small language models are likely far better than humans in this respect. 

We instead identify retention as the key challenge for memory in large language models (and probably more generally in deep learning models trained with gradient descent). We conjecture that improving memory retention in neural networks might alleviate some of the core problems afflicting modern large language models, such as their unreliability and their propensity to hallucinate. Ideas from the continual learning literature might be useful in this respect. On the other hand, perfect retention might be undesirable too from a generalization point of view (or from a novelty, creativity, or originality point of view). So, for many practically useful applications, there is likely a more productive middle ground between these two extremes.  

The memory retention curves we have found for our recall experiments (Figure~\ref{retention_fig}a, c) are qualitatively quite similar to the retention curves reported in many experimental studies on human long-term memory retention \citep{bahrick1975, bahrick1984, bahrick1987, conway1991, wixted1991, stanhope1993, rubin1996, custers2010}. These studies generally report a rapid decline in memory immediately after the learning episode (the induction of memory), followed by a more gradual decline, with \textit{some} residual memory remaining even after a very long interval without any explicit rehearsal. Our results are consistent with the suggestion that this type of retention curve might arise generically in memory systems under rather general conditions \citep{kahana2002}. 

Memory retention experiments with language models like those we have performed in this work can also serve as a valuable tool for applied cognitive psychologists by allowing them to formulate and test hypotheses about practically important ``cognitive design'' questions such as the optimal rehearsal schedule design for maximizing the long-term retention of acquired knowledge and to efficiently run very large-scale long-term memory retention experiments that would have been practically impossible to do with humans.

\bibliography{llm-memory}

\begin{thebibliography}{}

\bibitem[Bahrick, 1984]{bahrick1984}
Bahrick, H.~P. (1984).
\newblock {Semantic memory content in permastore: Fifty years of memory for
  Spanish learned in school.}
\newblock {\em Journal of Experimental Psychology: General}, 113(1):1.

\bibitem[Bahrick et~al., 1975]{bahrick1975}
Bahrick, H.~P., Bahrick, P.~O., and Wittlinger, R.~P. (1975).
\newblock Fifty years of memory for names and faces: A cross-sectional
  approach.
\newblock {\em Journal of Experimental Psychology: General}, 104(1):54.

\bibitem[Bahrick and Phelphs, 1987]{bahrick1987}
Bahrick, H.~P. and Phelphs, E. (1987).
\newblock {Retention of Spanish vocabulary over 8 years.}
\newblock {\em Journal of Experimental Psychology: Learning, Memory, and
  Cognition}, 13(2):344.

\bibitem[Brown et~al., 2020]{brown2020}
Brown, T., Mann, B., Ryder, N., Subbiah, M., Kaplan, J.~D., Dhariwal, P.,
  Neelakantan, A., Shyam, P., Sastry, G., Askell, A., et~al. (2020).
\newblock Language models are few-shot learners.
\newblock {\em Advances in Neural Information Processing Systems},
  33:1877--1901.

\bibitem[Carlini et~al., 2022]{carlini2022}
Carlini, N., Ippolito, D., Jagielski, M., Lee, K., Tramer, F., and Zhang, C.
  (2022).
\newblock Quantifying memorization across neural language models.
\newblock {\em arXiv preprint arXiv:2202.07646}.

\bibitem[Carlini et~al., 2019]{carlini2019}
Carlini, N., Liu, C., Erlingsson, {\'U}., Kos, J., and Song, D. (2019).
\newblock The secret sharer: Evaluating and testing unintended memorization in
  neural networks.
\newblock In {\em USENIX Security Symposium}, volume 267.

\bibitem[Carlini et~al., 2021]{carlini2021}
Carlini, N., Tramer, F., Wallace, E., Jagielski, M., Herbert-Voss, A., Lee, K.,
  Roberts, A., Brown, T.~B., Song, D., Erlingsson, U., et~al. (2021).
\newblock Extracting training data from large language models.
\newblock In {\em USENIX Security Symposium}, volume~6.

\bibitem[Conway et~al., 1991]{conway1991}
Conway, M.~A., Cohen, G., and Stanhope, N. (1991).
\newblock {On the very long-term retention of knowledge acquired through formal
  education: Twelve years of cognitive psychology.}
\newblock {\em Journal of Experimental Psychology: General}, 120(4):395.

\bibitem[Custers, 2010]{custers2010}
Custers, E.~J. (2010).
\newblock {Long-term retention of basic science knowledge: A review study}.
\newblock {\em Advances in Health Sciences Education}, 15(1):109--128.

\bibitem[Haviv et~al., 2022]{haviv2022}
Haviv, A., Cohen, I., Gidron, J., Schuster, R., Goldberg, Y., and Geva, M.
  (2022).
\newblock Understanding transformer memorization recall through idioms.
\newblock {\em arXiv preprint arXiv:2210.03588}.

\bibitem[Ippolito et~al., 2022]{ippolito2022}
Ippolito, D., Tram{\`e}r, F., Nasr, M., Zhang, C., Jagielski, M., Lee, K.,
  Choquette-Choo, C.~A., and Carlini, N. (2022).
\newblock Preventing verbatim memorization in language models gives a false
  sense of privacy.
\newblock {\em arXiv preprint arXiv:2210.17546}.

\bibitem[Kahana and Adler, 2017]{kahana2002}
Kahana, M.~J. and Adler, M. (2017).
\newblock Note on the power law of forgetting.
\newblock {\em {bioRxiv preprint, doi:10.1101/173765}}.

\bibitem[Kandpal et~al., 2022]{kandpal2022}
Kandpal, N., Deng, H., Roberts, A., Wallace, E., and Raffel, C. (2022).
\newblock Large language models struggle to learn long-tail knowledge.
\newblock {\em arXiv preprint arXiv:2211.08411}.

\bibitem[Kharitonov et~al., 2021]{kharitonov2021}
Kharitonov, E., Baroni, M., and Hupkes, D. (2021).
\newblock How {BPE} affects memorization in transformers.
\newblock {\em arXiv preprint arXiv:2110.02782}.

\bibitem[Lin, 2004]{lin2004}
Lin, C.-Y. (2004).
\newblock {ROUGE: A package for automatic evaluation of summaries}.
\newblock In {\em {Text Summarization Branches Out}}, pages 74--81.

\bibitem[Mallen et~al., 2022]{mallen2022}
Mallen, A., Asai, A., Zhong, V., Das, R., Hajishirzi, H., and Khashabi, D.
  (2022).
\newblock When not to trust language models: Investigating effectiveness and
  limitations of parametric and non-parametric memories.
\newblock {\em arXiv preprint arXiv:2212.10511}.

\bibitem[McCoy et~al., 2021]{mccoy2021}
McCoy, R.~T., Smolensky, P., Linzen, T., Gao, J., and Celikyilmaz, A. (2021).
\newblock How much do language models copy from their training data? evaluating
  linguistic novelty in text generation using {RAVEN}.
\newblock {\em arXiv preprint arXiv:2111.09509}.

\bibitem[Ouyang et~al., 2022]{ouyang2022}
Ouyang, L., Wu, J., Jiang, X., Almeida, D., Wainwright, C., Mishkin, P., Zhang,
  C., Agarwal, S., Slama, K., Ray, A., et~al. (2022).
\newblock Training language models to follow instructions with human feedback.
\newblock {\em {Advances in Neural Information Processing Systems}},
  35:27730--27744.

\bibitem[Rubin and Wenzel, 1996]{rubin1996}
Rubin, D.~C. and Wenzel, A.~E. (1996).
\newblock {One hundred years of forgetting: A quantitative description of
  retention.}
\newblock {\em Psychological Review}, 103(4):734.

\bibitem[Shepard, 1967]{shepard1967}
Shepard, R.~N. (1967).
\newblock Recognition memory for words, sentences, and pictures.
\newblock {\em Journal of Verbal Learning and Verbal Behavior}, 6(1):156--163.

\bibitem[Stanhope et~al., 1993]{stanhope1993}
Stanhope, N., Cohen, G., and Conway, M. (1993).
\newblock Very long-term retention of a novel.
\newblock {\em Applied Cognitive Psychology}, 7(3):239--256.

\bibitem[Tirumala et~al., 2022]{tirumala2022}
Tirumala, K., Markosyan, A.~H., Zettlemoyer, L., and Aghajanyan, A. (2022).
\newblock Memorization without overfitting: Analyzing the training dynamics of
  large language models.
\newblock {\em arXiv preprint arXiv:2205.10770}.

\bibitem[Wixted and Ebbesen, 1991]{wixted1991}
Wixted, J.~T. and Ebbesen, E.~B. (1991).
\newblock On the form of forgetting.
\newblock {\em Psychological Science}, 2(6):409--415.

\bibitem[Zhang et~al., 2021]{zhang2021}
Zhang, C., Ippolito, D., Lee, K., Jagielski, M., Tram{\`e}r, F., and Carlini,
  N. (2021).
\newblock Counterfactual memorization in neural language models.
\newblock {\em arXiv preprint arXiv:2112.12938}.

\end{thebibliography}
\bibliographystyle{apalike}

\end{document}